\documentclass[10pt,twocolumn,letterpaper]{article}

\usepackage[pagenumbers]{cvpr} 

\definecolor{cvprblue}{rgb}{0.21,0.49,0.74}
\usepackage[pagebackref,breaklinks,colorlinks,allcolors=cvprblue]{hyperref}
\usepackage{multirow}
\usepackage[accsupp]{axessibility}
\def\paperID{} 
\def\confName{CVPR}
\def\confYear{2026}

\title{Parallax to Align Them All: An OmniParallax Attention Mechanism for Distributed Multi-View Image Compression}

\author{
Haotian Zhang$^1$ \quad Feiyue Long$^1$ \quad Yixin Yu$^1$ \quad Jian Xue$^1$ \quad Haocheng Tang$^1$ \\
Tongda Xu$^2$ \quad Zhenning Shi$^3$\quad Yan Wang$^2$ \quad Siwei Ma$^1$  \quad Jiaqi Zhang$^1$\thanks{Corresponding author.}      \\
$^1$The National Engineering Laboratory for Video Technology, School of Computer Science, Peking University\\
$^2$Institute for AI Industry Research (AIR), Tsinghua University \\
$^3$Tsinghua Shenzhen International Graduate School, Tsinghua University
}



\begin{document}

\maketitle
\begin{abstract}
Multi-view image compression (MIC) aims to achieve high compression efficiency by exploiting inter-image correlations, playing a crucial role in 3D applications. As a subfield of MIC, distributed multi-view image compression (DMIC) offers performance comparable to MIC while eliminating the need for inter-view information at the encoder side.
However, existing methods in DMIC typically treat all images equally, overlooking the varying degrees of correlation between different views during decoding, which leads to suboptimal coding performance. 
To address this limitation, we propose a novel $\textbf{OmniParallax Attention Mechanism}$ (OPAM), which is a general mechanism for explicitly modeling correlations and aligned features between arbitrary pairs of information sources.
Building upon OPAM, we propose a Parallax Multi Information Fusion Module (PMIFM) to adaptively integrate information from different sources. PMIFM is incorporated into both the joint decoder and the entropy model to construct our end-to-end DMIC framework, $\textbf{ParaHydra}$.
Extensive experiments demonstrate that $\textbf{ParaHydra}$ is $\textbf{the first DMIC method}$ to significantly surpass state-of-the-art MIC codecs, while maintaining low computational overhead. Performance gains become more pronounced as the number of input views increases. Compared with LDMIC, $\textbf{ParaHydra}$ achieves bitrate savings of $\textbf{19.72\%}$ on WildTrack(3) and up to $\textbf{24.18\%}$ on WildTrack(6), while significantly improving coding efficiency (as much as $\textbf{65}\times$ in decoding and $\textbf{34}\times$ in encoding).
\end{abstract}    
\begin{figure}[t]
\centering
\includegraphics[width=0.5\textwidth]
{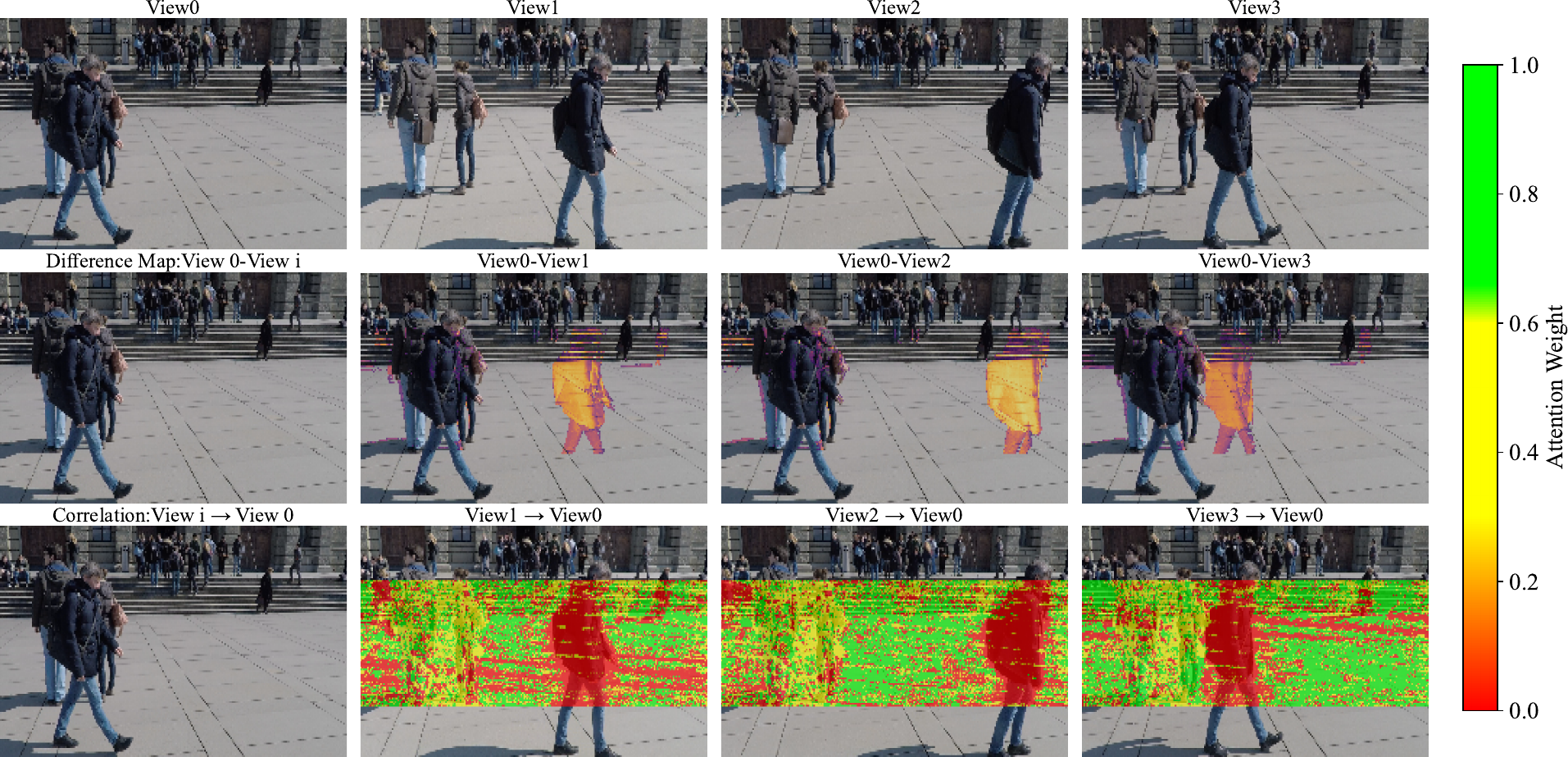} 
\caption{
\textbf{Visualization of OPAM correlations.} \textbf{Row 1}: Input views. \textbf{Row 2}: Difference maps. \textbf{Row 3}: Correlations, where consistent regions ({\color{green}{{green}}}) are prioritized over occlusions ({\color{red}{{red}}}).
}
\label{fig:attention}
\end{figure} 

\section{Introduction}

Multi-view image compression (MIC)~\cite{jabbireddy2024vemic, zhang2023ldmic,huang20243d} aims to achieve high compression efficiency by jointly compressing multiple views and exploiting inter-image correlations. It plays a crucial role in applications such as autonomous driving~\cite{yin20203d}, virtual reality~\cite{jin2021subjective}, and robot navigation~\cite{sanchez2018survey}. Recent work~\cite{huang20243d} proposes a learning-based multi-view image coding approach that incorporates 3D Gaussian geometric priors, achieving superior compression performance. However, these methods rely on strong prior knowledge of different views during encoding, which may not be available in many real-world scenarios. Motivated by distributed source coding (DSC) theory \cite{slepian2003noiseless,wyner2003rate}, which shows that separate encoding and joint decoding of correlated sources can match the compression performance of joint encoding-decoding, researchers have explored distributed multi-view image compression (DMIC)~\cite{zhang2023ldmic, mital2023neural,huang2023learned}. In this paradigm, multi-view images are independently encoded and transmitted, then jointly reconstructed at the decoder using information from all viewpoints. This approach eliminates the need for prior view knowledge during encoding, while maintaining competitive compression performance, making it particularly suitable for practical multi-camera applications.

LDMIC \cite{zhang2023ldmic}, the first end-to-end DMIC framework, achieves comparable coding performance to MIC methods. However, its use of average pooling for fusing side-view features is suboptimal, as it assigns equal importance to all views regardless of their semantic relevance. As illustrated in Fig.~\ref{fig:attention}, when reconstructing regions such as floor in the main view, it is preferable to exploit side views where the floor is clearly visible and unoccluded. Treating all side views equally may introduce noise (e.g., occlusions like pedestrians), which degrades reconstruction quality. To effectively leverage multi-view information, accurately measuring inter-source correlations remains a central challenge.

Inspired by the parallax attention mechanism (PAM) used in stereo matching~\cite{zhang2024stereo,wang2020parallax,wang2021symmetric}, which identifies cross-view correspondences by computing attention between two positions along the same row of left and right views, we explore the use of PAM to measure inter-source correlations. However, PAM restricts attention computation to positions along the same epipolar line, thereby limiting its ability to aggregate information. To address this limitation, we present a comprehensive derivation and propose a novel $\textbf{OmniParallax Attention Mechanism}$ (OPAM), a general mechanism for explicitly modeling correlations and aligned features between arbitrary pairs of information sources. OPAM efficiently captures the full two-dimensional (2D) spatial context with cubic computational complexity, offering a significantly more efficient alternative to full 2D self-attention, which incurs quartic complexity.

Building on OPAM, we propose a Parallax Multi Information Fusion Module (PMIFM), which is a general multi-source feature integration module guided by the correlations provided by OPAM. Leveraging PMIFM, we further propose two core components: the Parallax Joint Decoder (Para-JD), which integrates features from multiple views, and the Parallax Entropy Model (Para-EM), which aggregates channel-wise, local spatial, and global spatial contexts. Together, these modules form our end-to-end DMIC framework, \textbf{ParaHydra}, which supports compression and reconstruction for an \textbf{arbitrary} number of input views.

Extensive experiments demonstrate that \textbf{ParaHydra} is \textbf{the first DMIC method} to significantly surpass state-of-the-art (SOTA) MIC codecs, while maintaining excellent scalability.
Compared with the SOTA MIC codec LMVIC, \textbf{ParaHydra} achieves a bitrate saving of up to \textbf{34.11\%} on Mip-NeRF 360(4). Compared with LDMIC, \textbf{ParaHydra} achieves a bitrate saving of \textbf{24.18\%} on WildTrack(6), while significantly improving coding efficiency (as much as $\textbf{65}\times$ in decoding and $\textbf{34}\times$ in encoding). Our main contributions are summarized as follows:
\begin{itemize}
    \item We present a detailed and rigorous derivation of PAM and propose the novel OPAM, which is a general mechanism for explicitly modeling correlations and aligned features between arbitrary pairs of information sources. OPAM efficiently captures the full two-dimensional spatial context with cubic computational complexity.
    \item Building upon OPAM, we propose PMIFM, a general multi-source feature integration module. PMIFM is incorporated into both joint decoder and entropy model to form our end-to-end DMIC framework, \textbf{ParaHydra}.
    \item \textbf{ParaHydra} exhibits excellent scalability, enabling compression and reconstruction for an \textbf{arbitrary} number of input views while maintaining stable runtime.
    \item Extensive experimental results show that \textbf{ParaHydra} is \textbf{the first DMIC method} to significantly surpass SOTA MIC codecs, while maintaining low computational overhead. Performance gains become more pronounced as the number of input views increases.

\end{itemize}

\section{Related Work}
\subsection{Single Image Compression}
Traditional codecs \cite{marcellin2000overview, bross2021overview} adopt hand-crafted modular pipelines for transform, quantization, and entropy coding, limiting joint optimization and overall performance. Deep learning-based end-to-end frameworks \cite{balle2016end} have since become dominant, with advanced entropy models significantly improving rate-distortion (RD) efficiency \cite{balle2018variational, minnen2018joint}. The checkerboard context model (CCM) \cite{he2021checkerboard} achieves a good trade-off between RD performance and complexity via a two-pass anchor/non-anchor design, while MLIC \cite{jiang2023mlic++} further integrates local, global, and channel-wise contexts. However, existing methods still underutilize latent representations. Our approach introduces a more effective entropy model that better captures latent dependencies for superior compression performance.
\subsection{Multi-View Image Compression}
Research on multi-view image compression has primarily focused on stereo settings with two views, commonly referred to as stereo image compression (SIC). Existing methods can be broadly categorized into unidirectional prediction-based~\cite{liu2019dsic, wodlinger2022sasic} and bidirectional attention-based~\cite{liu2024bidirectional, wodlinger2024ecsic, lei2022deep}. MSFDPM~\cite{huang2023learned} performs feature matching based on Pearson correlation, whereas the proposed OPAM explicitly models both correlations and aligned features through efficient attention mechanisms. Although research on SIC has made considerable progress, compression for scenarios involving three or more views remains underexplored. Traditional approaches such as MV-HEVC~\cite{vetro2011overview} rely on hand-crafted disparity prediction, while learning-based methods like LMVIC~\cite{huang20243d} heavily depend on strong prior knowledge of cross-view relationships during encoding, which limits their applicability in real-world scenarios. LDMIC~\cite{zhang2023ldmic} is the first to introduce the DSC paradigm into MIC, achieving impressive performance. However, its use of simple average pooling neglects the semantic relationships among views. In contrast, our method models semantic relevance more precisely, enabling efficient multi-view fusion and superior reconstruction quality.

\begin{figure*}[t] 
  \centering
  \subfloat[ ]{
    \label{subfig:parahydra}
\includegraphics[width=0.5\textwidth,height=0.2\textheight]
    {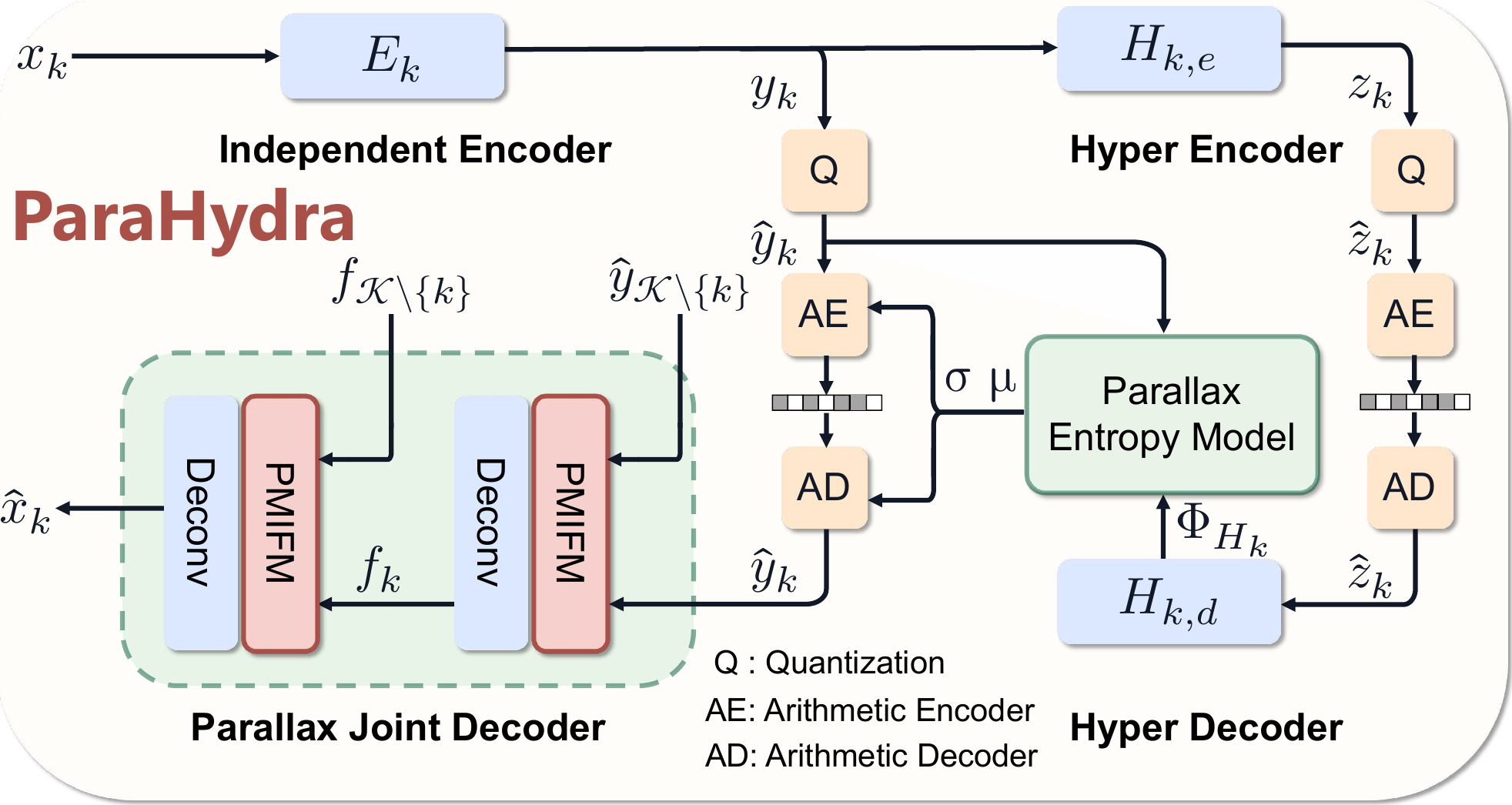}
  }
  \subfloat[ ]{
    \label{subfig:pmifm}
\includegraphics[width=0.18\textwidth,height=0.2\textheight]
{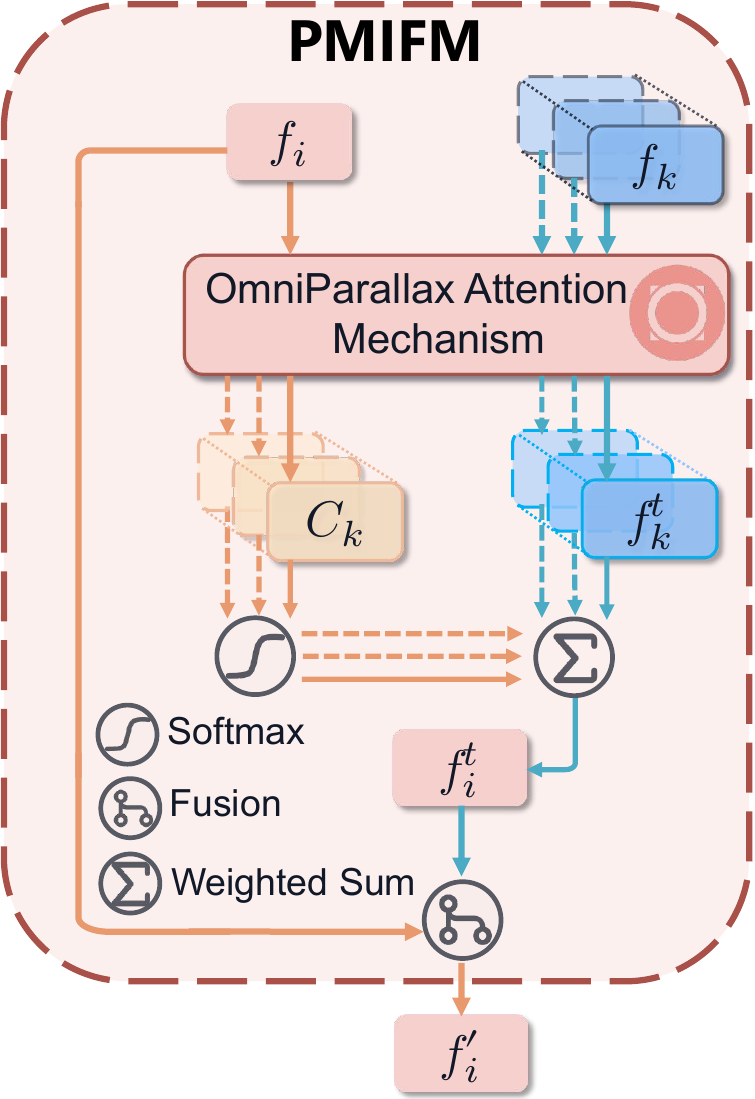}
  }
  \subfloat[ ]{
    \label{subfig:para-EM}
\includegraphics[width=0.29\textwidth,height=0.2\textheight]
{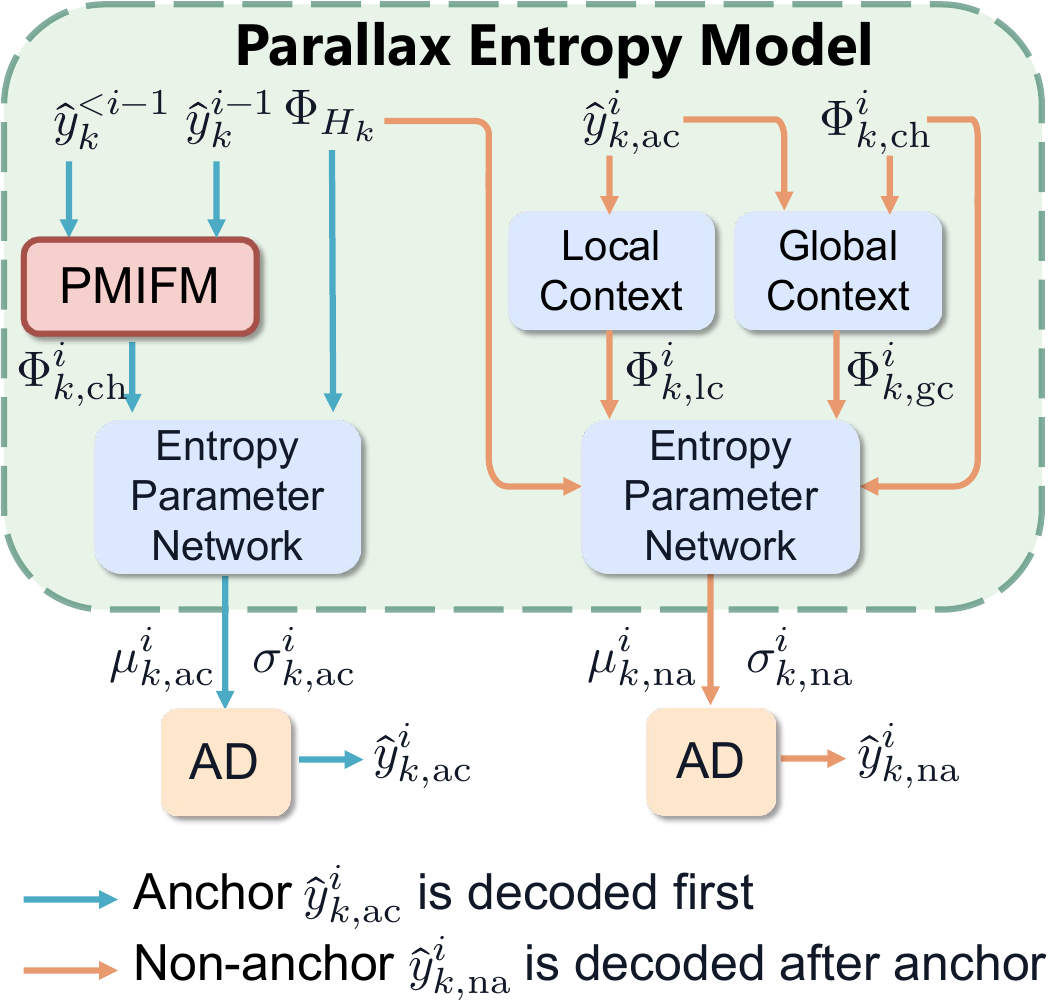}
  }
\caption{
\textbf{(a) The proposed ParaHydra framework.} $\hat{y}_{\mathcal{K}\setminus \{k\}}$ and ${f}_{\mathcal{K}\setminus \{k\}}$ denote the sets of all view features excluding the $k$-th view feature ${\hat{y}}_k$ and ${f}_k$, respectively. 
\textbf{(b) Parallax Multi Information Fusion Module (PMIFM).} PMIFM integrates the aligned feature $f_k^{t}$ based on the semantic relevance $C_k$ between each side source $f_k$ and the main source $f_i$. 
\textbf{(c) Parallax Entropy Model.} The figure illustrates the decoding process for the slice $\hat{y}^i_k$. Each latent slice $\hat{y}^i_k$ is partitioned into anchor $\hat{y}^i_{k,\text{ac}}$ and non-anchor $\hat{y}^i_{k,\text{na}}$ parts. Anchor $\hat{y}^i_{k,\text{ac}}$ is decoded first using Gaussian parameters $(\mu^i_{k,\text{ac}}, \sigma^i_{k,\text{ac}})$ predicted from the channel context $\Phi^i_{k,\text{ch}}$ (provided by previous slices $\hat{y}^{<i}_k$) and the hyperprior $\Phi_{H_k}$. Non-anchor $\hat{y}^i_{k,\text{na}}$ is decoded next using Gaussian parameters $(\mu^i_{k,\text{na}}, \sigma^i_{k,\text{na}})$ predicted from the local context $\Phi^i_{k,\text{lc}}$ (derived from anchor $\hat{y}^i_{k,\text{ac}}$), the channel context $\Phi^i_{k,\text{ch}}$, the global context $\Phi^i_{k,\text{gc}}$, and the hyperprior $\Phi_{H_k}$.}
  \label{figParaHydra}
\end{figure*}
\section{Proposed Method}
\subsection{Overview}

An overview of the proposed {ParaHydra} framework is illustrated in Fig.~\ref{subfig:parahydra}. Let $\mathcal{K} = \{1, 2, \dots, K\}$ denote the index set of $K$ multi-view images. Given a group of images $x_{\mathcal{K}} = \{x_1, x_2, \dots, x_K\}$, each image $x_k$ is independently encoded into a latent representation $y_k$ using its dedicated encoder $E_k$. The latent feature $y_k$ is then transformed into a hyper-latent $z_k$ via a hyper encoder module $H_{k,e}$:
\begin{equation}
y_k = E_k(x_k), \quad z_k = H_{k,e}(y_k),  \quad\forall k \in \mathcal{K}.
\end{equation}

Both $y_k$ and $z_k$ are then quantized to obtain $\hat{y}_k$ and $\hat{z}_k$, the quantized hyper-latent $\hat{z}_k$ is entropy-coded using a fully factorized prior model $\psi$:
\begin{equation}
R(\hat{z}_k) = \mathbb{E} \left[ -\log \left( p_{\hat{z} \mid \psi} (\hat{z}_k \mid \psi) \right) \right],
\end{equation}
where $\mathbb{E}[\cdot]$ denotes the expectation. For the quantized latent representation $\hat{y}_k$, we employ a slice-wise checkerboard-based entropy model. Specifically, $\hat{y}_k$ is divided into $l$ channel slices: $\hat{y}_k = [\hat{y}^1_k, \hat{y}^2_k, \dots, \hat{y}^l_k]$, each slice $\hat{y}^i_k$ is partitioned into anchor $ \hat{y}^i_{k,\text{ac}}$ and non-anchor $\hat{y}^i_{k,\text{na}}$ parts, modeled by Gaussian entropy models:
\begin{equation}
\begin{aligned}
R(\hat{y}^i_{k,\text{ac}}) &= \mathbb{E} \left[ -\log p_{\hat{y}^i_{k,\text{ac}}}(\hat{y}^i_{k,\text{ac}}) \right],  \hat{y}^i_{k,\text{ac}} \sim \mathcal{N}(\mu^i_{k,\text{ac}}, \sigma^i_{k,\text{ac}}), \\
R(\hat{y}^i_{k,\text{na}}) &= \mathbb{E} \left[ -\log p_{\hat{y}^i_{k,\text{na}}}(\hat{y}^i_{k,\text{na}}) \right],  \hat{y}^i_{k,\text{na}} \sim \mathcal{N}(\mu^i_{k,\text{na}}, \sigma^i_{k,\text{na}}),
\end{aligned}
\end{equation}
where $\mathcal{N}(\mu, \sigma)$ denotes the Gaussian distribution. The Gaussian parameters are predicted by the Entropy Parameter (EP) network, conditioned on multiple feature sources:
\begin{equation}
\begin{aligned}
\mu^i_{k,\text{ac}}, \sigma^i_{k,\text{ac}} &= EP(\Phi^i_{k,\text{ch}}, \Phi_{H_k}), \\
\mu^i_{k,\text{na}}, \sigma^i_{k,\text{na}} &= EP(\Phi^i_{k,\text{ch}}, \Phi_{H_k}, \Phi^i_{k,\text{lc}}, \Phi^i_{k,\text{gc}}),
\end{aligned}
\end{equation}
here, $\Phi_{H_k}$ is the hyperprior feature extracted via hyper decoder $H_{k,d}$. $\Phi^i_{k,\text{ch}}$, $\Phi^i_{k,\text{lc}}$, and $\Phi^i_{k,\text{gc}}$ represent the channel-wise, local, and global contexts, respectively, all provided by the Para-EM.
Once all quantized latent representations $\hat{y}_{\mathcal{K}}=\{\hat{y}_1, \hat{y}_2, \dots,\hat{y}_K\}$ are obtained, the Para-JD aggregates inter-view features to reconstruct the image set:
\begin{equation}
\label{eq:para_jd}
\hat{x}_{\mathcal{K}} = \text{Para-JD}(\hat{y}_{\mathcal{K}}).
\end{equation}

ParaHydra is optimized end-to-end based on the RD loss function:
\begin{equation}
L = \lambda D + R = \lambda \sum_{k=1}^{K} d(x_k, \hat{x}_k) + \sum_{k=1}^{K} \big( R(\hat{y}_k) + R(\hat{z}_k) \big),
\end{equation}
where $d(x_k, \hat{x}_k)$ denotes the distortion between $x_k$ and $\hat{x}_k$, $R(\hat{y}_k)$ and $R(\hat{z}_k)$ are the estimated compression rates of the latent and hyper-latent representations, and $\lambda$ controls the trade-off between rate and distortion. 

In the following sections, we first describe the proposed OPAM in Section~\ref{subsec:opam}. Building upon OPAM, we introduce PMIFM in Section~\ref{subsec:pmifm}. PMIFM is further integrated into the Para-JD (Section~\ref{subsec:para-jd}) and Para-EM (Section~\ref{subsec:para-em}) modules.

\subsection{OmniParallax Attention Mechanism}
\label{subsec:opam}
\begin{figure}[t] 
\centering
\includegraphics[width=\columnwidth]
{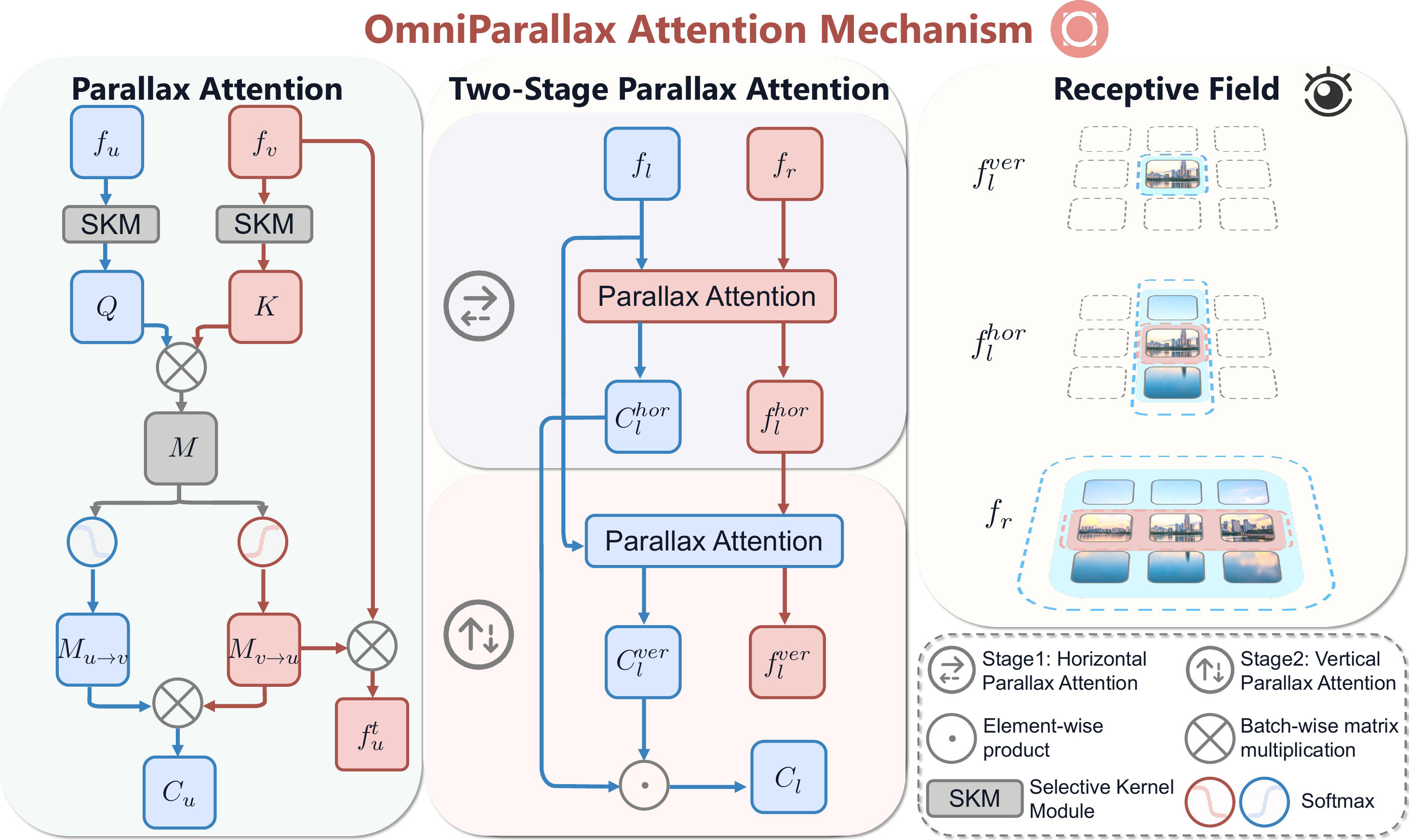} 
\caption{
\textbf{Overview of OmniParallax Attention Mechanism (OPAM).}
\textit{Left:} \textbf{Parallax attention.} 
\textit{Middle:} \textbf{Two-stage parallax attention in OPAM.} OPAM applies horizontal (red) and vertical (blue) parallax attention sequentially to capture the full 2D spatial context. 
\textit{Right:} \textbf{Receptive fields of the aligned features.} Each position in $f_l^{{hor}}$ attends to one row of $f_r$, and each position in $f_l^{{ver}}$ attends to one column of $f_l^{{hor}}$, allowing each position in $f_l^{{ver}}$ to attend to the entire 2D spatial domain of $f_r$.
}
\label{figOPAM}
\end{figure}

In LDMIC \cite{zhang2023ldmic}, average pooling is used to integrate information from side views. However, this strategy is suboptimal, as it assigns equal importance to all views regardless of their semantic relevance. To overcome this limitation, we propose the OmniParallax Attention Mechanism (OPAM), which efficiently explores the full two-dimensional spatial context of the side information source to provide a reliable reference and corresponding consistency for the main source. The consistency reflects the reliability of the reference and can be interpreted as the semantic relevance between the main and side sources.
OPAM consists of two complementary components: Horizontal Parallax Attention (HPA) and Vertical Parallax Attention (VPA). The key difference is that HPA performs attention along the horizontal axis, while VPA operates along the vertical axis. An overview of the OPAM is shown in Fig.~\ref{figOPAM}. In this section, we first present the general formulation of parallax attention, which underlies both HPA and VPA. We then detail the two-stage parallax attention process in OPAM, where HPA and VPA are applied sequentially to capture full two-dimensional spatial context. Detailed mathematical formulations and analysis of OPAM are provided in the Appendix.
\subsubsection{Parallax Attention}
In this subsection, we take HPA as an example to illustrate the process of the parallax attention mechanism. The VPA follows the same procedure, except that the reshape operations applied to the row dimension in HPA are instead applied to the column dimension in VPA.

Let the main and side information sources be denoted by $f_u,f_v \in \mathbb{R}^{B\times H\times W\times C}$. We first compute the query feature map $Q \in \mathbb{R}^{B \times H \times W \times C}$ and key feature map $K  \in \mathbb{R}^{B \times H \times W \times C}$ using a Selective Kernel Module (SKM)~\cite{li2019selective}:
\begin{equation}
    Q = \text{SKM}(f_u), \quad K = \text{SKM}(f_v),
\end{equation}
the row dimension of $Q$ and $K$ is reshaped into the batch dimension to obtain $Q_b, K_b \in \mathbb{R}^{BH \times W \times C}$. The cross-correlation map $M \in \mathbb{R}^{BH \times W \times W}$ is then computed as:
\begin{equation}
    M = Q_b \otimes K_b^{\top},
\end{equation}
where $\otimes$ denotes batch-wise matrix multiplication and $K_b^{\top}$ represents the transpose of $K_b$ over its last two dimensions.

The parallax attention maps, denoted as $M_{v \to u}$,$ M_{u \to v}\in \mathbb{R}^{BH \times W \times W}$, are obtained by applying softmax function to $M$ and its transpose $M^{\top}$ along the last dimension:
\begin{equation}
\begin{aligned}
M_{v \to u} &= \text{softmax}(M,\ \text{dim}=-1), \\
M_{u \to v} &= \text{softmax}(M^{\top},\ \text{dim}=-1).
\end{aligned}
\end{equation}

Next, the second dimension of $M_{v \to u}$ and the third dimension of $M_{u \to v}$ are reshaped into the batch dimension to obtain $M_{v \to u,b} \in \mathbb{R}^{BHW \times 1 \times W}$ and $M_{u \to v,b} \in \mathbb{R}^{BHW \times W \times 1}$. These are then used to compute the batch cycle consistency \cite{wang2020parallax} $C_{u,b} \in \mathbb{R}^{BHW \times 1 \times 1}$:
\begin{equation}
\label{eq_V}
C_{u,b} = M_{v \to u,b} \otimes M_{u \to v,b},
\end{equation}
the resulting $C_{u,b}$ is then reshaped to obtain the cycle consistency $C_u \in \mathbb{R}^{B \times H \times W}$.
Finally, the row dimension of $f_v \in \mathbb{R}^{B \times H \times W \times C}$ is reshaped into the batch dimension to obtain $f_{v,b} \in \mathbb{R}^{BH \times W \times C}$. The batch aligned feature $f_{u,b}^t \in \mathbb{R}^{BH \times W \times C}$ is then computed as:
\begin{equation}
\label{eq_hor}
f_{u,b}^t = M_{v \to u} \otimes f_{v,b},
\end{equation}
and $f_{u,b}^t$ is further reshaped to form the final aligned feature $f_{u}^{t} \in \mathbb{R}^{B \times H \times W \times C}$.
Overall, the outputs of PAM are the aligned feature $f_u^{t}$ and the corresponding consistency $C_u$. The aligned feature $f_u^{t}$ serves as a reference derived from $f_v$, where each position in $f_u^{t}$ aggregates information from all positions along the corresponding row of $f_v$. Meanwhile, each position in $C_u$ indicates the reliability of the corresponding location in $f_u^{t}$.
The computational complexity of PAM is $O(N^3)$, where $N = \max(H, W)$.
\subsubsection{Two-Stage Parallax Attention in OPAM}
Given the main source $f_l$ and side source $f_r$, the HPA is first applied to capture dependencies along the row dimension, producing the horizontally aligned feature $f_l^{hor}$ and the horizontal cycle consistency $C_l^{hor}$. Next, $f_l$ and $f_l^{hor}$ are passed into the VPA, which models vertical dependencies and produces the vertically aligned feature $f_{l}^{ver}$ along with the vertical cycle consistency $C_{l}^{ver}$.

The overall cycle consistency $C_l$ is computed by combining both horizontal and vertical consistency:
\begin{equation}
\label{eq:c_hor_ver}
    C_l = C_{l}^{hor} \odot C_{l}^{ver},
\end{equation}
where $\odot$ denotes the element-wise product. 

By combining Eq.~\ref{eq_hor}, $f_{l}^{ver}$ can be expressed as:
 \begin{equation}
 \begin{split}
 \label{eq:f_hor_ver}
 &f_{l}^{ver}[b, i, j, k] 
 = \sum_{g=0}^{H-1} M_{r \to l}^{ver}[bW + j, i, g]   f_{l}^{hor}[b, g, j, k] =\\
 &\sum_{g=0}^{H-1} \sum_{s=0}^{W-1}  M_{r \to l}^{ver}[bW + j, i, g]   M_{r \to l}^{hor}[bH + g, j, s] f_r[b,g,s,k],
 \end{split}
 \end{equation}
where $M_{r \to l}^{hor}$ denotes the horizontal parallax attention map from $f_r$ to $f_l$, and $M_{r \to l}^{ver}$ denotes the vertical parallax attention map from $f_l^{hor}$ to $f_l$. This formulation shows that by sequentially applying horizontal and vertical parallax attention, the attention is no longer restricted to aggregation along a single epipolar line. Instead, it can exploit the full two-dimensional spatial context (Fig.~\ref{figOPAM}\textit{ Right}), providing a reliable reference $f_{l}^{ver}$ and corresponding consistency $C_l$. 
The consistency $C_l$ captures the joint reliability of context information provided by $f_r$ for reconstructing $f_l$, guided by both horizontal and vertical parallax attention. It can be interpreted as the semantic relevance between $f_r$ and $f_l$.

Since the complexity of one stage of parallax attention is $O(N^3)$, the overall complexity of OPAM remains $O(N^3)$, which is far more efficient than full two-dimensional self-attention with complexity $O(N^4)$.

\subsection{Parallax Multi Information Fusion Module}
\label{subsec:pmifm}
To adaptively fuse information from multiple sources, we propose the Parallax Multi Information Fusion Module (PMIFM), as illustrated in Fig.~\ref{subfig:pmifm}. PMIFM integrates features based on the semantic relevance between each side source $f_k$ and the main source $f_i$. The overall formulation of PMIFM is defined as:
\begin{equation}
f_i^{\prime} = \text{PMIFM}(f_i, f_{\mathcal{S}}),
\end{equation}
where $f_{\mathcal{S}}$ denotes the set of side sources, and $f_i^{\prime}$ represents the refined representation of $f_i$, enriched through the adaptive fusion of side information.

For each side source $f_k$ ($k \ne i$), OPAM is applied to $f_i$ and $f_k$ to obtain the aligned feature $f_k^{t}$ and its corresponding consistency $C_k$. These consistency maps are concatenated to form a global consistency vector: $C = [C_1, \dots, C_{i-1}, C_{i+1}, \dots]$. A softmax function is then applied to $C$ to compute normalized attention 
weights: $W = \text{softmax}(C)$. Using the attention weights $W$, the relevance-weighted fused feature $f_i^{t}$ is computed as:
\begin{equation}
f_i^{t} = \sum_{k \ne i} W_k \cdot f_k^{t}.
\end{equation}

Finally, a lightweight fusion network $F$~\cite{zhang2023ldmic} integrates $f_i^{t}$ with the original input $f_i$:
\begin{equation}
f_i^{\prime} = F(f_i^{t}, f_i).
\end{equation}

\subsection{Parallax Joint Decoder}
\label{subsec:para-jd}
The proposed Parallax Joint Decoder (Para-JD) aggregates inter-view features to reconstruct the image set. An illustration of the Para-JD architecture is provided in Fig.~\ref{subfig:parahydra}, and the overall process is formulated in Eq.~\ref{eq:para_jd}. Para-JD is composed of two submodules, each consisting of a PMIFM followed by a deconvolutional network. The quantized latent representations $\hat{y}_{\mathcal{K}}$ are first input into the first submodule of Para-JD to produce the latent feature set $f_{\mathcal{K}} = \{f_1, f_2, \dots, f_K\}$. For each view $k$, the representation $\hat{y}_k$ is refined using PMIFM, which adaptively integrates features from all other views $\hat{y}_{\mathcal{K} \setminus \{k\}}$ to generate latent feature $f_k$:
\begin{equation}
f_k = \text{Deconv}\big(\text{PMIFM}(\hat{y}_k, \hat{y}_{\mathcal{K} \setminus \{k\}})\big),
\end{equation}
the resulting latent features $f_{\mathcal{K}}$ are subsequently passed through the second submodule of Para-JD to reconstruct the image set $\hat{x}_{\mathcal{K}}$, where each $f_k$ is further refined by PMIFM using information from $f_{\mathcal{K} \setminus \{k\}}$.

\subsection{Parallax Entropy Model}
\begin{figure*}[t]
  \centering
  \includegraphics[width=0.25\textwidth]{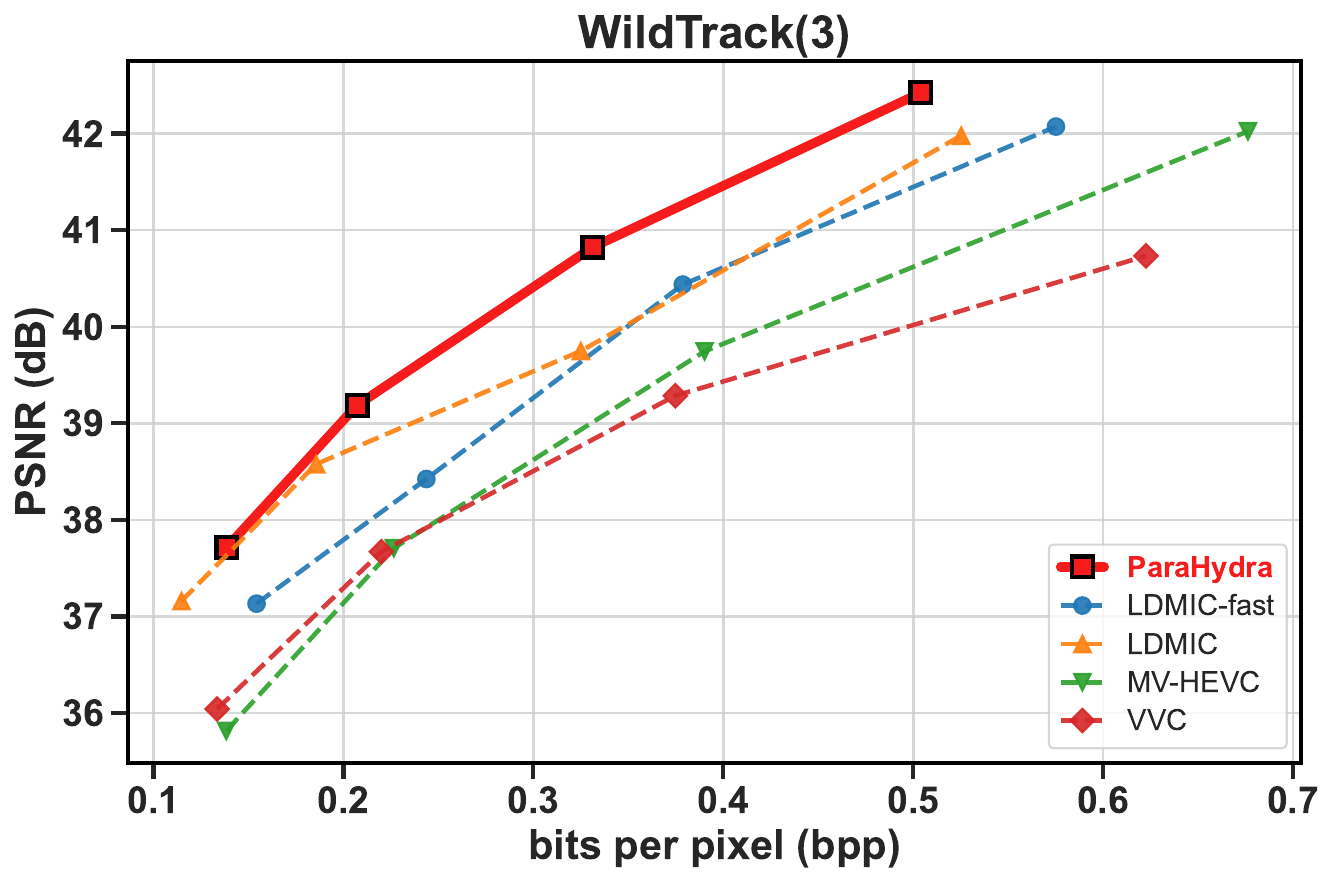}\hfill
  \includegraphics[width=0.25\textwidth]{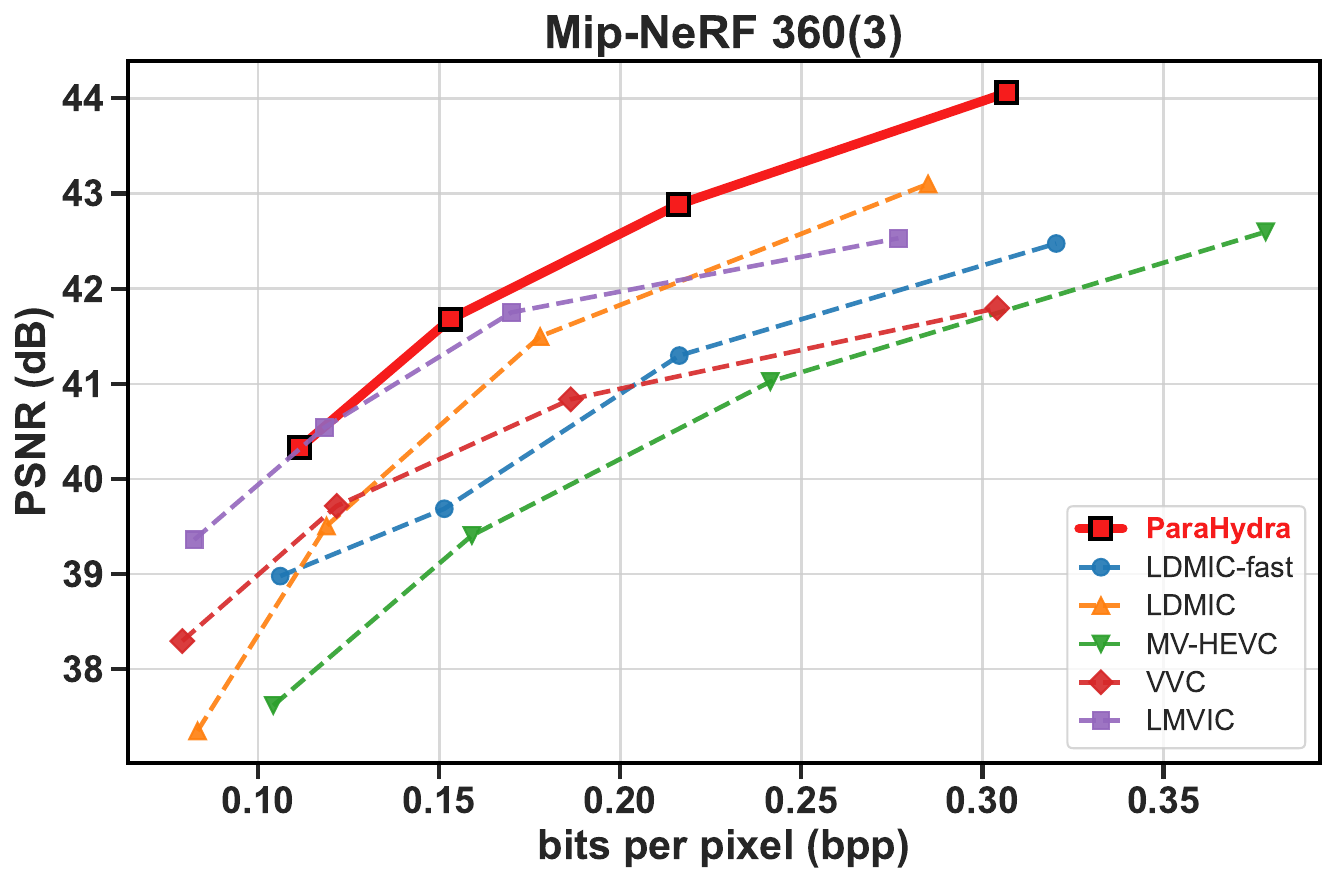}\hfill
    \includegraphics[width=0.25\textwidth]{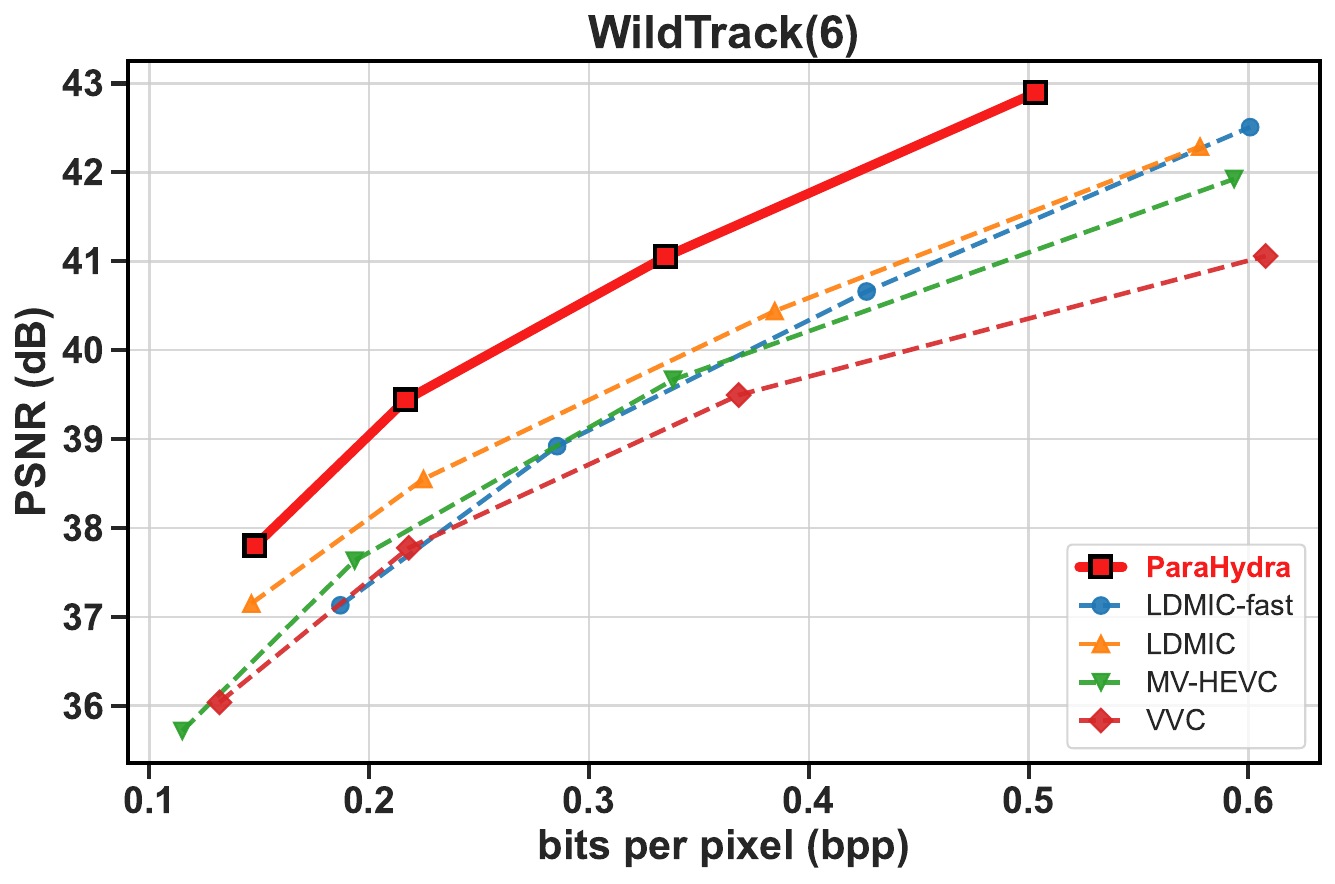}\hfill
  \includegraphics[width=0.25\textwidth]{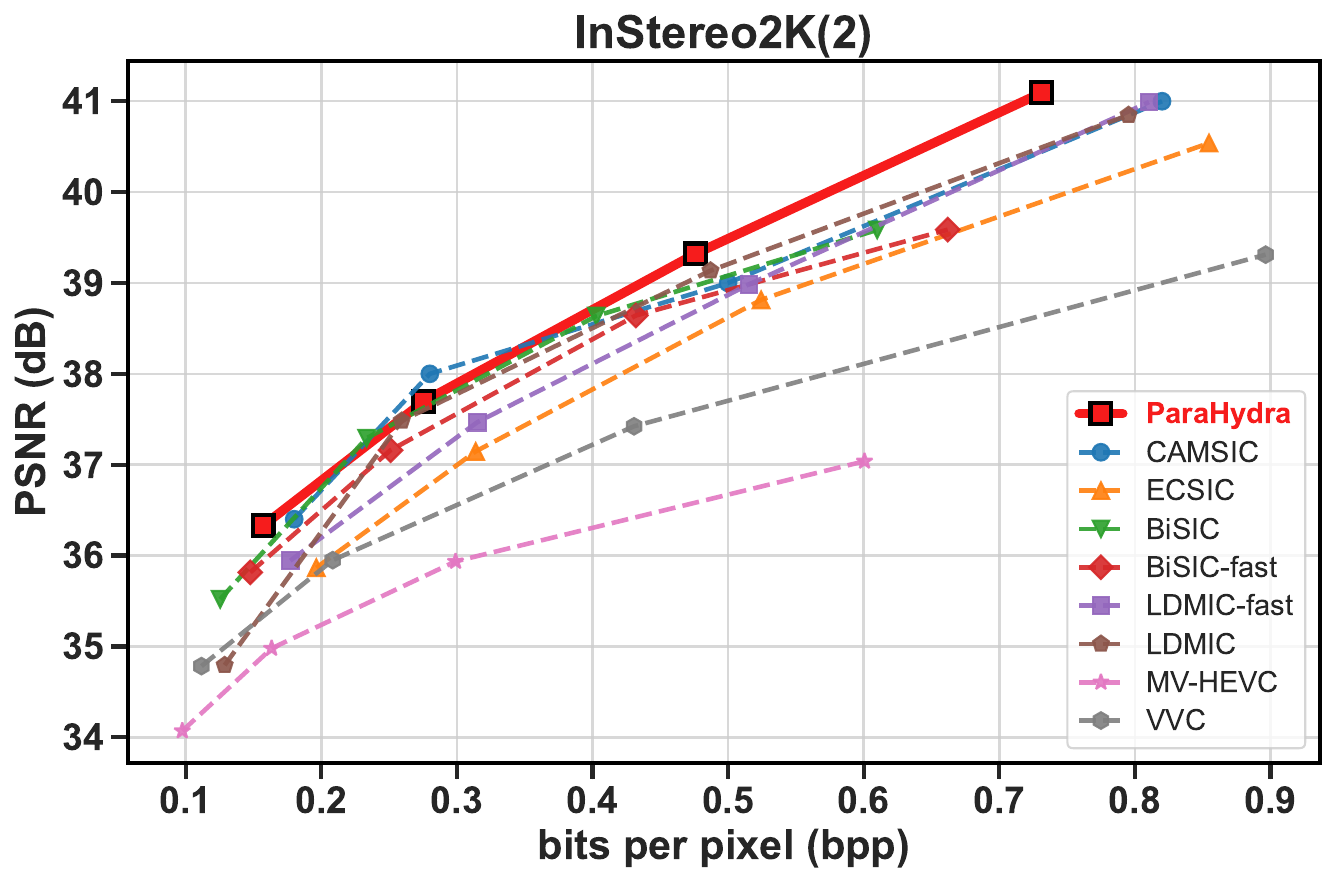}
  \caption{
  \textbf{Rate-distortion curves of ParaHydra compared against various baselines.}
  }
  \label{fig:rd_performance}
\end{figure*}
\label{subsec:para-em}
MLIC \cite{jiang2023mlic++} effectively captures multi-scale spatial and channel-wise contexts, yet it overlooks certain correlations within latent representations, leaving room for further improvement. To address this limitation, we design the Parallax Entropy Model (Para-EM), which leverages PMIFM to measure correlations within latent representations and aggregate context information more effectively.

The overall architecture of Para-EM is shown in Fig.~\ref{subfig:para-EM}. Para-EM consists of three components: the Parallax Channel Context Module, the checkerboard attention-based local context module~\cite{jiang2023mlic++}, and the Parallax Global Context Module, which are introduced in the following sections.

\subsubsection{Parallax Channel Context Module}
LDMIC \cite{zhang2023ldmic} employs an auto-regressive context model to capture spatial dependencies in $\hat{y}$, but this incurs high computational cost and overlooks channel correlations in the 3D latent space. MLIC \cite{jiang2023mlic++} introduces a channel-wise context module that concatenates previous channels for decoding, yet it treats all channels equally, allowing less informative ones to introduce noise and degrade context representation. 

To address this limitation and fully exploit channel context, we propose a Parallax Channel Context Module (PCCM), which uses PMIFM to adaptively aggregate channel information and provide more effective context. Let the quantized latent representation be denoted as $\hat{y}_k \in \mathbb{R}^{C \times H \times W}$. We partition $\hat{y}_k$ along the channel dimension into non-overlapping slices: $\hat{y}_k = [\hat{y}_k^1, \hat{y}_k^2, \dots, \hat{y}_k^l]$, where $l$ denotes the total number of channel slices, and each slice $\hat{y}_k^i \in \mathbb{R}^{s_c \times H \times W}$, $s_c$ is the channel number of each slice. The slices are processed sequentially.
When decoding the $i$-th slice $\hat{y}_k^i$, we use the previously decoded slices $\hat{y}_k^{<i}$ to provide channel-wise contextual information. In this setting, we reinterpret “information source” in OPAM as a channel slice. Specifically, the most recently decoded slice $\hat{y}_k^{i-1}$ is treated as the reference, while all earlier slices $\hat{y}_k^{<i-1}$ are aggregated using PMIFM to produce channel context $\Phi_{k,\text{ch}}^i$:
\begin{equation}
    \Phi_{k,\text{ch}}^i = \text{PMIFM}(\hat{y}_k^{i-1}, \hat{y}_k^{< i - 1}).
\end{equation}
\subsubsection{Parallax Global Context Module}
The intra-slice global context module ($g_{\text{gc,intra}}$)~\cite{jiang2023mlic++} estimates correlations between the anchor $\hat{y}^{i}_{k,\text{ac}}$ and non-anchor $\hat{y}^{i}_{k,\text{na}}$ parts within a latent slice $\hat{y}^{i}_k$. However, this estimation is limited because it relies solely on global patterns extracted from the immediately preceding slice $\hat{y}^{i-1}_k$, thereby overlooking potentially informative contextual cues from earlier slices $\hat{y}^{<i-1}_k$. To overcome this limitation and enhance intra-slice correlation modeling, we propose a Parallax Global Context Module (PGCM). PGCM exploits all previously decoded slices $\hat{y}^{<i}_k$ to construct a more comprehensive global context $\Phi_{k,\text{gc}}^{i}$ for the current slice.

First, PCCM is applied to extract global, channel-aware features $\Phi_{k,\text{ch}}^{i}$ from all earlier slices $\hat{y}^{<i}_k$. Next, the global intra-slice attention mechanism $g_{\text{gc,intra}}$ is used to estimate correlations between anchor and non-anchor parts in $\hat{y}^i_k$, conditioned on the enhanced channel context $\Phi_{k,\text{ch}}^{i}$:
\begin{equation}
\begin{aligned}
\hat{y}^{i}_{k,\text{attn}} &= \text{softmax} \!\left( \frac{Q^{<i}_{k,\text{na}}\times (K^{<i}_{k,\text{ac}})^{\top}}{\sqrt{s_c}} \right) \times V^{i}_{k,\text{ac}}, \\[4pt]
\hat{y}^{i}_{k,\text{conv}} &= \text{Conv}(\hat{y}^{i}_{k,\text{attn}}), \\[4pt]
\Phi_{k,\text{gc}}^{i} &= \text{DepthRB}(\hat{y}^{i}_{k,\text{conv}}),
\end{aligned}
\end{equation}
where $V^{i}_{k,\text{ac}}$ denotes the value projection of the anchor $\hat{y}^i_{k,\text{ac}}$, while $Q^{<i}_{k,\text{na}}$ and $K^{<i}_{k,\text{ac}}$ represent the query and key projections derived from the non-anchor and anchor parts of the channel context $\Phi_{k,\text{ch}}^{i}$, respectively. $\text{DepthRB}(\cdot)$ denotes the depth-wise residual bottleneck block \cite{jiang2023mlic++}.

\section{Experiments}
\begin{table*}[t]
\caption{  \textbf{BDBR comparison relative to LDMIC across different datasets.} The best results are highlighted in {\color{purple}{\textbf{bold}}}. The number of input views is indicated in parentheses. A dash (\textendash)  denotes that the corresponding model cannot be evaluated on that dataset.}
\label{table:performance_table}
\centering
\resizebox{\textwidth}{!}{
\begin{tabular}{l*{5}{cc}}
\toprule
\multirow{2}{*}{Methods} &
\multicolumn{2}{c}{InStereo2K(2)} &
\multicolumn{2}{c}{Cityscapes(2)} &
\multicolumn{2}{c}{WildTrack(3)} &
\multicolumn{2}{c}{WildTrack(6)} &
\multicolumn{2}{c}{Mip-NeRF 360(3)} \\
\cmidrule(lr){2-3} \cmidrule(lr){4-5} \cmidrule(lr){6-7} \cmidrule(lr){8-9} \cmidrule(lr){10-11}
& PSNR & MS-SSIM & PSNR & MS-SSIM & PSNR & MS-SSIM & PSNR & MS-SSIM & PSNR & MS-SSIM \\
\midrule
LDMIC &0\% &0\%  & 0\% & 0\% & 0\% & 0\% &0\%  & 0\% &0\%  & 0\% \\
VVC &48.68\% &80.69\%  & 54.33\% & 101.92\% & 49.47\% & 115.64\% &25.16\%  & 120.67\% &7.14\%  & 96.47\% \\
MV-HEVC & 84.84\%  &182.11\%  & 106.27\% & 202.95\% & 31.84\% & 176.86\% & 10.01\% & 168.04\% & 41.15\% &171.15\%
  \\
ECSIC &25.65\%   & 11.99\% & 50.51\% & 41.12\% & \textendash & \textendash & \textendash & \textendash & \textendash & \textendash \\
BiSIC & -2.95\%  &-4.65\%  & 11.40\% & -7.11\% & \textendash & \textendash & \textendash & \textendash & \textendash & \textendash \\
BiSIC-fast & 7.37\%  &-3.09\%  & 18.65\% & -7.00\% & \textendash & \textendash & \textendash & \textendash & \textendash & \textendash \\
CAMSIC &1.05\%  & 4.98\% & 8.64\% & -5.02\% & \textendash & \textendash & \textendash & \textendash & \textendash & \textendash \\
LMVIC & \textendash & \textendash & \textendash & \textendash & \textendash & \textendash & \textendash & \textendash & -14.30\% & 93.73\% \\
LDMIC-fast & 13.15\%  &-1.85\%  & 3.90\% & 24.58\% & 10.83\% & 9.24\% &10.51\% &9.67\%  &25.52\%  & 16.05\% \\
\textbf{ParaHydra} & \color{purple}{\textbf{-6.92\%}}   & \color{purple}{\textbf{-5.95\%}} & \color{purple}{\textbf{-1.42\%}}  &  \color{purple}{\textbf{-7.15\%}} & \color{purple}{\textbf{-19.72\%}}  & \color{purple}{\textbf{-7.07\%}}  &  \color{purple}{\textbf{-24.18\%}} & \color{purple}{\textbf{-6.23\%}}  & \color{purple}{\textbf{-18.20\%}}  & \color{purple}{\textbf{-4.53\%}} \\
\bottomrule
\end{tabular}}
\end{table*}

\subsection{Experimental Setup}
\textbf{Baselines.} We compare our method with several SOTA baselines, including the traditional codecs VVC \cite{bross2021overview} and MV-HEVC \cite{vetro2011overview}; SIC methods ECSIC \cite{wodlinger2024ecsic}, BiSIC, BiSIC-fast \cite{liu2024bidirectional}, and CAMSIC \cite{zhang2025camsic}; MIC method LMVIC \cite{huang20243d};  and DMIC methods LDMIC and LDMIC-fast \cite{zhang2023ldmic}. We reimplement all methods to ensure a fair comparison.\\
\textbf{Datasets.} To evaluate the coding performance of the proposed framework, we adopt two widely used stereo datasets, InStereo2K~\cite{bao2020instereo2k} and Cityscapes~\cite{cordts2016cityscapes}, as well as two multi-view datasets, WildTrack~\cite{chavdarova2018wildtrack} and Mip-NeRF 360~\cite{barron2022mip}. These datasets cover a broad range of indoor and outdoor scenes, each containing thousands of images captured from different viewpoints, with resolutions ranging from $1080 \times 860$ to over 4K. Training and testing settings on the datasets follow previous works~\cite{zhang2023ldmic, liu2024bidirectional} to ensure a fair comparison.\\
\textbf{Metrics.} Reconstruction quality is measured by the Peak Signal-to-Noise Ratio (PSNR) and the Multi-Scale Structural Similarity Index (MS-SSIM). The bitrate is reported in bits per pixel (bpp). In addition to plotting rate-distortion (RD) curves, we compute the Bjøntegaard Delta Bitrate (BDBR) to quantify average bitrate savings across different quality levels and the Bjøntegaard Delta Quality (BD-PSNR) to measure the quality gains~\cite{bjontegaard2001calculation, barman2024bjontegaard}.

\subsection{Implementation Details}
The implementation follows that of~\cite{zhang2023ldmic, jiang2023mlic++, li2019selective}. Specifically, the number of channels for the latent representations $y_k$, $z_k$, and $f_k$ is set to 192. The number of channel slices $l$ is set to 8, resulting in $s_c=24$ channels per slice. The window size $w$ is set to 5. All learning-based models are trained with the trade-off parameter $\lambda = 1024, 2048, 4096,8192\, (32, 64, 128,256)$ under MSE (MS-SSIM). We train our models for 1400 epochs on multi-view datasets with a batch size of 8 and for 3000 epochs on stereo image datasets with a batch size of 16, using a learning rate of $10^{-4}$. Experiments are conducted on a single NVIDIA A30 GPU. More details are provided in the Appendix.
\begin{table*}[t]
   \caption{ \textbf{Complexity comparison of various codecs evaluated on images with a resolution of 1024$\times$832 from the InStereo2K dataset.}} 
  \label{table:codec_runtime}
  \centering
  \scriptsize 
  \resizebox{\linewidth}{!}{
  \begin{tabular}{ccccccccc}
    \toprule
    & \multicolumn{8}{c}{Methods} \\
    \cmidrule(lr){2-9}
    & VVC & MV-HEVC  & BiSIC-fast & CAMSIC & BiSIC & LDMIC & \textbf{2D Attn} & \textbf{ParaHydra} \\
    \midrule
    Enc. Time (s) & 349.47 & 24.67  & 0.35 & 0.87 & 0.30 & 9.27 & 0.62 & 0.27 \\
    Dec. Time (s) & 0.37 & 0.17  & 0.41 & 1.79 & 0.49 & 21.43 & 0.67 & 0.33 \\
    Params (M) & -- & --  & 327.66 & 602.05 & 305.71 & 214.98 & 108.81 & 105.25 \\
    FLOPs (T) & -- & --  & 3.96 & 9.18 & 3.77 & 2.88 & 1.64 & 1.78 \\
    \bottomrule
  \end{tabular}
  }
\end{table*}
\begin{table}[t]
\centering
\caption{ \textbf{BDBR comparison relative to LDMIC on Mip-NeRF.}}
\label{table:performance_Mip_small}
\resizebox{\columnwidth}{!}{%
\begin{tabular}{cccccc}
\toprule
Views & VVC & MV-HEVC & LDMIC-fast & LMVIC & \textbf{ParaHydra} \\
\midrule
3 & 7.14\% & 41.15\% & 25.52\% & -14.30\% & {\color{purple}\textbf{-18.20\%}} \\
4 & 53.06\% & 91.30\% & 14.31\% & 17.27\% & {\color{purple}\textbf{-16.84\%}} \\
\bottomrule
\end{tabular}%
}
\end{table}

\begin{figure}[t]
  \centering
  \includegraphics[scale=0.3]{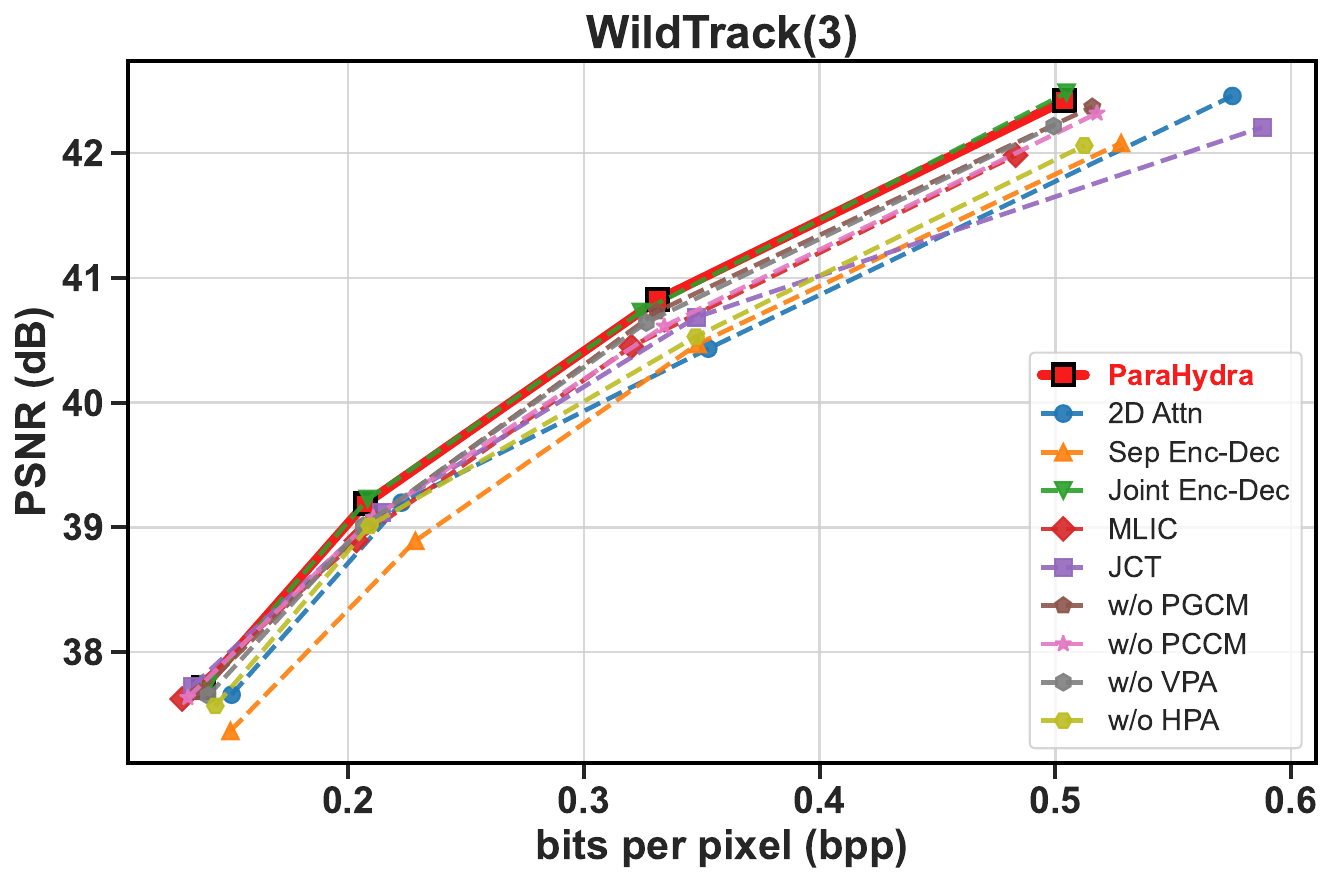}
  \caption{
  \textbf{Ablation study on the WildTrack with 3 input views.} 
  }

  \label{fig:ablation}

 \end{figure}
 
\subsection{Quantitative Results}
\textbf{RD Performance.} Fig.~\ref{fig:rd_performance} shows the RD curves of all compared methods based on PSNR, while RD curves based on MS-SSIM are provided in the Appendix. Tab.~\ref{table:performance_table} reports the BDBR results of each codec, measured relative to LDMIC. Across all datasets, ParaHydra consistently outperforms baseline codecs in both PSNR and MS-SSIM. \textbf{The performance gains become more pronounced as the number of input views increases.} When the number of input views is limited to two, the main view can reference only a single side view, which restricts multi-view fusion capacity of PMIFM. Despite this limitation, ParaHydra still outperforms SOTA SIC methods by a clear margin. This result highlights ParaHydra’s excellent capability for efficient information utilization.
On multi-view datasets with three input views, ParaHydra achieves average bitrate savings of 19.72\% and 18.20\% on WildTrack and Mip-NeRF 360, respectively, compared to LDMIC. When the number of views increases to six, the average bitrate saving further improves to \textbf{24.18\%} on WildTrack. These results demonstrate that, in complex multi-view scenarios, OPAM effectively captures semantic correlations across views, thereby enhancing inter-view information utilization. Notably, on the 4K-resolution Mip-NeRF 360 dataset, ParaHydra surpasses SOTA MIC codec LMVIC, which leverages 3D Gaussian geometric priors during encoding. As shown in Tab.~\ref{table:performance_Mip_small}, with three input views, ParaHydra achieves an average bitrate saving of 3.9\% compared with LMVIC. When the number of views increases to four, the average bitrate saving further improves to an impressive \textbf{34.11\%}. This finding highlights not only the robustness of ParaHydra in high-resolution scenarios but also the capability of the proposed DMIC framework to outperform joint encoding-decoding methods by efficiently exploiting inter-view information.\\
\textbf{Computational Complexity.} Tab.~\ref{table:codec_runtime} summarizes computational complexity of various codecs. ParaHydra achieves the best compression performance while maintaining low computational overhead, being up to \textbf{65}$\times$ faster in decoding and \textbf{34}$\times$ faster in encoding compared to LDMIC. This efficiency stems from DMIC paradigm and checkerboard-based entropy model, which together enable parallel processing in both encoder and entropy coding stages. 
\subsection{Ablation Study}
\textbf{Ablation Settings.} We conduct ablation study on the WildTrack dataset with three input views to evaluate the contribution of each module. All models are optimized for MSE, and the RD curves are shown in Fig.~\ref{fig:ablation}. The results reported below are compared against our ParaHydra.\\
\textbf{Inter-view Information Utilization.} To assess the contributions of the proposed modules in fusing inter-view information, we control whether encoder and decoder have access to inter-view context. At the same bitrate level, PSNR of the model with both encoder and decoder access (denoted as Joint Enc-Dec) increases by only approximately \textbf{0.01~dB}, while the model with neither access (Sep Enc-Dec) decreases by as much as \textbf{0.54~dB}. These results indicate that with PMIFM, an efficient information fusion module, it is sufficient for only decoder to access multi-view information to achieve performance comparable to joint encoding-decoding. We further replace OPAM with either JCT operation \cite{zhang2023ldmic} (JCT) or 2D self-attention (2D Attn), which increases the bitrate by 5.64\% or 11.30\%, respectively. 
These demonstrate that OPAM effectively captures semantic correlations across different viewpoints, thereby enabling more efficient utilization of multi-view information. 
Removing HPA (w/o HPA) or VPA (w/o VPA) in OPAM increases the bitrate by 8.79\% or 3.63\%, respectively. 
Either HPA or VPA restricts attention computation within a single epipolar line, limiting the ability to aggregate information. In contrast, OPAM achieves more effective information aggregation by fully exploring 2D spatial context. 
As shown in Tab.~\ref{table:codec_runtime}, 2D Attn introduces significant computational overhead, about 230\% encoding time and 203\% decoding time compared with OPAM, highlighting the superior efficiency of OPAM. \\
\textbf{Entropy Model.} To evaluate the effectiveness of the proposed Para-EM in exploiting multiple contexts within entropy models, we individually replace the proposed channel-wise and global context modules with their counterparts from~\cite{jiang2023mlic++}. The results indicate that replacing either the channel-wise module (w/o PCCM) or the global context module (w/o PGCM) increases the bitrate by 4.32\% or 2.92\%, respectively. Furthermore, replacing Para-EM with the MLIC module~\cite{jiang2023mlic++} (MLIC) increases the bitrate by 5.13\%. These results validate the effectiveness of the proposed context modules and further highlight the versatility of the proposed PMIFM, which not only facilitates the fusion of information across multiple views but also enables efficient aggregation of channel information.
\begin{figure}[t]
    \centering
    \newcommand{\figheight}{6cm}
    \captionsetup[subfigure]{aboveskip=0pt, belowskip=0pt}
    \begin{minipage}[t]{0.12\textwidth} 
        \centering
        \subcaption*{Original Image}
        \includegraphics[height=3.6\figheight]{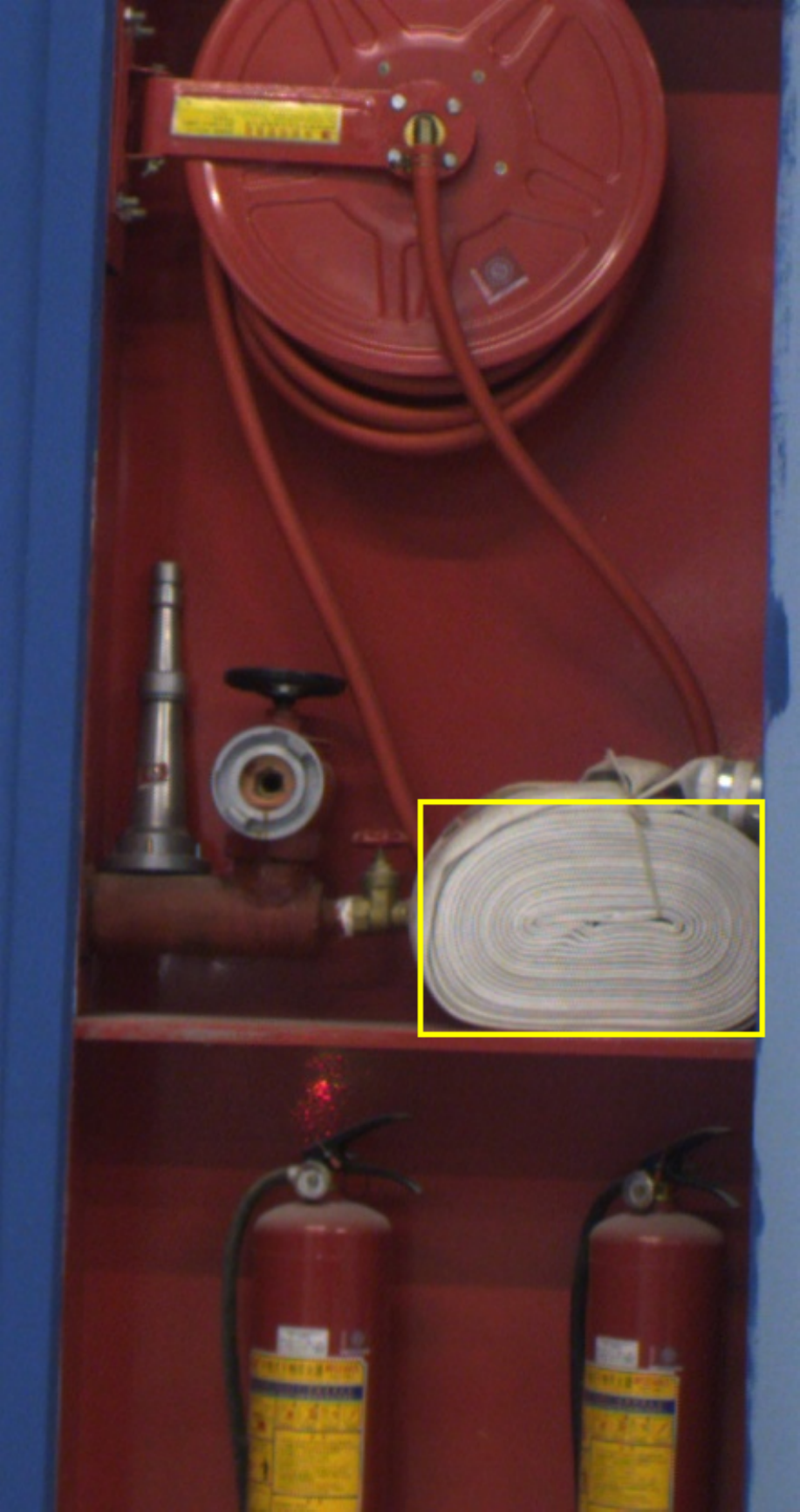}
    \end{minipage}%
    \begin{minipage}[t]{0.379\textwidth}


        \renewcommand{\arraystretch}{0} 
        \newcommand{\modelname}[1]{\makebox[1.9cm][c]{#1}}

        \begin{subfigure}[t]{0.3\textwidth}
        \centering
            \subcaption*{\modelname{Ground truth}}
            \includegraphics[height=1.4\figheight]{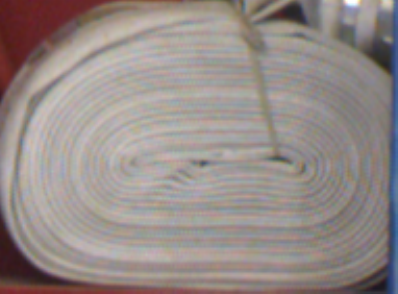}
            \subcaption*{PSNR / bpp}    
        \end{subfigure}%
        \begin{subfigure}[t]{0.31\textwidth}
        \centering
            \subcaption*{\modelname{CAMSIC}}
    \includegraphics[height=1.4\figheight]{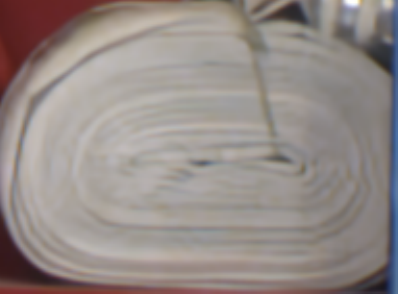}
            \subcaption*{36.27 / 0.10}
        \end{subfigure}%
        \begin{subfigure}[t]{0.3\textwidth}
        \centering
            \subcaption*{\modelname{BiSIC-fast}}
            \includegraphics[height=1.4\figheight]{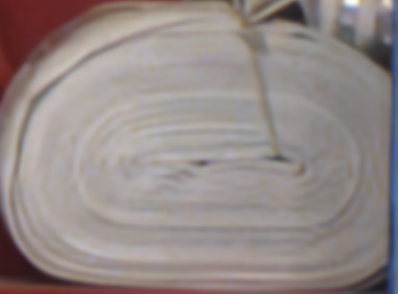}
            \subcaption*{36.11 / 0.11}
        \end{subfigure}

        \begin{subfigure}[t]{0.286\textwidth}
        \centering
            \subcaption*{LDMIC-fast}
            \includegraphics[height=1.4\figheight]{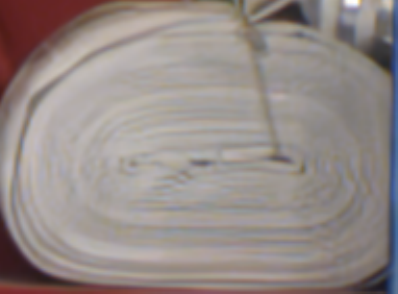}
            \subcaption*{36.68 / 0.09}
        \end{subfigure} %
        \begin{subfigure}[t]{0.311\textwidth}
        \centering
        \subcaption*{\modelname{LDMIC}}
            \includegraphics[height=1.4\figheight]{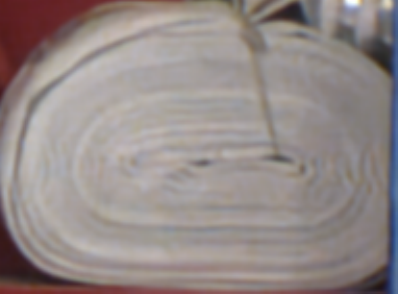}
            \subcaption*{36.25 / 0.09}
        \end{subfigure}%
        \begin{subfigure}[t]{0.29\textwidth}
        \centering
        \subcaption*{\modelname{\textbf{Ours}}}
            \includegraphics[height=1.4\figheight]{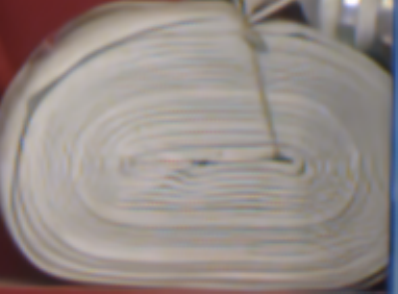}
            \subcaption*{
            {\color{purple}\textbf{37.04}} / {\color{purple}\textbf{0.08}}}
            
        \end{subfigure}\\[0pt]
        
    \end{minipage}

    \caption{\textbf{Qualitative results on the InStereo2K.} PSNR and bitrate (bpp) are denoted as PSNR / bpp. Zoom in for a better view.}
    \label{fig:Qualitative results}
\end{figure}
\subsection{Qualitative Analysis}
Fig.~\ref{fig:Qualitative results} presents qualitative results on the InStereo2K dataset. ParaHydra achieves the highest compression ratio while maintaining the best reconstruction quality. When the bitrate falls below 0.1~bpp, other models exhibit noticeable distortions in fine structural regions, such as hydrant, whereas ParaHydra consistently preserves fine-grained details. These observations demonstrate the robustness and effectiveness of Para-EM in leveraging context information. Even at extremely low bitrates, Para-EM remains capable of effectively utilizing context information within latent representations to achieve high-quality image reconstruction. More qualitative results are provided in the Appendix. 

To further analyze the proposed OPAM, Fig.~\ref{fig:attention} visualizes the inter-view correlations generated by OPAM. The first row shows the input views, while the second row displays the difference maps between the main and side views, explicitly highlighting occlusions. These maps align with the OPAM correlations (Row 3), confirming OPAM accurately suppresses  these occluded regions ({\color{red}{{red}}}) and prioritizes consistent regions ({\color{green}{{green}}}). This highlights OPAM's \textbf{excellent capability in capturing high-level semantic correlations}.

\section{Conclusion}
This paper presents ParaHydra, \textbf{the first DMIC method} to significantly surpass SOTA MIC codecs, while maintaining excellent scalability. To mitigate the limitation of neglecting semantic relevance among information sources in prior works, we propose the novel OmniParallax Attention Mechanism (OPAM), a general mechanism for explicitly modeling correlations and aligned features between arbitrary pairs of information sources. OPAM efficiently captures the full 2D spatial context with cubic computational complexity. Building on OPAM, we propose Parallax Multi Information Fusion Module to adaptively fuse multi-source information, which is further employed in the joint decoder and entropy model of ParaHydra. 
Extensive experiments demonstrate that ParaHydra outperforms existing methods by a significant margin while maintaining low computational overhead. Notably, performance gains of ParaHydra become increasingly pronounced as the number of input views scales up.
\newpage
\section*{Acknowledgements} 
This work was supported in part by the Beijing Natural Science Foundation (Grant No.~L242014), in part by the National Natural Science Foundation of China (Grant No.~62501022), in part by the R24115SG MIGU-PKU Meta Vision Technology Innovation Lab, and in part by the New Cornerstone Science Foundation through the XPLORER Prize. It was also supported by the Wuxi Research Institute of Applied Technologies, Tsinghua University (Grant No.~20242001120). 

{
    \small
    \bibliographystyle{ieeenat_fullname}
    \bibliography{main}
}

\end{document}